\documentclass{vgtc}                          % final (conference style)
\ifpdf%                                % if we use pdflatex
  \pdfoutput=1\relax                   % create PDFs from pdfLaTeX
  \usepackage{graphicx}                % allow us to embed graphics files
  \DeclareGraphicsExtensions{.pdf,.png,.jpg,.jpeg} % for pdflatex we expect .pdf, .png, or .jpg files
\else%                                 % else we use pure latex
  \usepackage{graphicx}                % allow us to embed graphics files
  \DeclareGraphicsExtensions{.eps}     % for pure latex we expect eps files
\fi%

\graphicspath{{figures/}{pictures/}{images/}{./}} % where to search for the images

\usepackage{microtype}                 % use micro-typography (slightly more compact, better to read)
\PassOptionsToPackage{warn}{textcomp}  % to address font issues with \textrightarrow
\usepackage{textcomp}                  % use better special symbols
\usepackage{mathptmx}                  % use matching math font
\usepackage{times}                     % we use Times as the main font
\usepackage{cite}                      % needed to automatically sort the references
\usepackage{tabu}                      % only used for the table example
\usepackage{booktabs}                  % only used for the table example
\usepackage{multirow}
\usepackage{xspace}
\newcommand{\toolname}{Latent Space Explorer\xspace}
\usepackage{enumitem}

\onlineid{1148}

\vgtccategory{Research}

\vgtcinsertpkg

\title{\toolname: Visual Analytics \\for Multimodal Latent Space Exploration}

\author{Bum Chul Kwon\thanks{e-mail: bumchul.kwon@us.ibm.com}\\ %
        \scriptsize IBM Research %
\and Samuel Friedman\thanks{e-mail: sam@broadinstitute.org}\\ %
     \scriptsize Broad Institute of MIT and Harvard %
\and Kai Xu\thanks{e-mail: Kai.Xu@nottingham.ac.uk}\\ %
     \scriptsize University of Nottingham %
\and Steven A Lubitz\thanks{e-mail: lubitz@broadinstitute.org}\\ %
     \scriptsize Broad Institute of MIT and Harvard
\and Anthony Philippakis\thanks{e-mail: aphilipp@broadinstitute.org}\\ %
     \scriptsize Broad Institute of MIT and Harvard
\and Puneet Batra\thanks{e-mail: pbatra@broadinstitute.org}\\ %
     \scriptsize Broad Institute of MIT and Harvard
\and Patrick T Ellinor\thanks{e-mail: pellinor@broadinstitute.org}\\ %
     \scriptsize Broad Institute of MIT and Harvard
\and Kenney Ng\thanks{e-mail: kenney.ng@us.ibm.com}\\ %
        \scriptsize IBM Research}

\teaser{
  \centering
  \includegraphics[width=\linewidth]{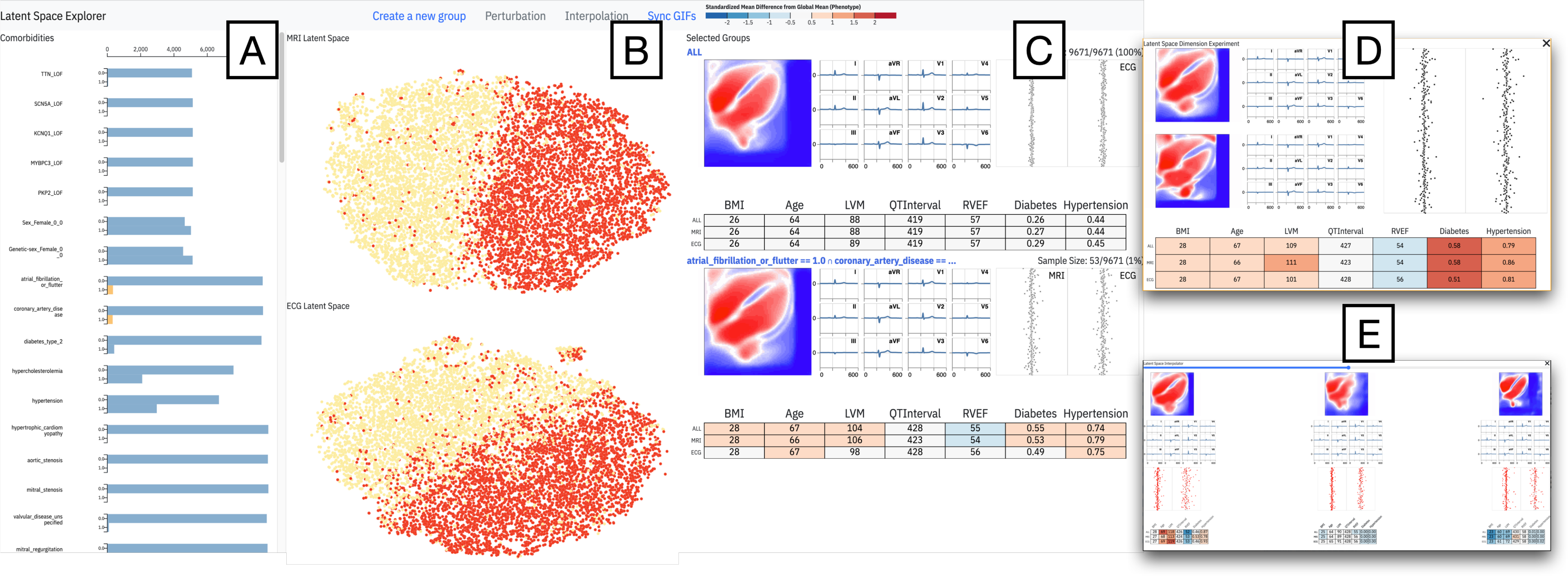}
  \caption{Overview of \toolname with the example of cardiac MRIs and ECGs: (A) users can view the distribution over demographic variables and phenotypes; (B) users can explore groups of subjects with respect to similarities in latent space of MRIs (top) and ECGs (bottom); (C) users create and inspect user-defined groups of subjects in details; (D) users can interactively reconstruct MRIs and ECGs by sampling new instances with perturbed latent space feature values; (E) In addition, users can interactively interpolate samples between two groups of subjects in latent spaces of MRIs and ECGs.}
  \label{fig:teaser}
}

\abstract{Machine learning models built on training data with multiple modalities can reveal new insights that are not accessible through unimodal datasets. 
For example, cardiac magnetic resonance images (MRIs) and electrocardiograms (ECGs) are both known to capture useful information about subjects' cardiovascular health status. 
A multimodal machine learning model trained from large datasets can potentially predict the onset of heart-related diseases and provide novel medical insights about the cardiovascular system. 
Despite the potential benefits, it is difficult for medical experts to explore multimodal representation models without visual aids and to test the predictive performance of the models on various subpopulations.
To address the challenges, we developed a visual analytics system called \textit{\toolname}. 
\toolname provides interactive visualizations that enable users to explore the multimodal representation of subjects, define subgroups of interest, interactively decode data with different modalities with the selected subjects, and inspect the accuracy of the embedding in downstream prediction tasks. 
A user study was conducted with medical experts and their feedback provided useful insights into how \toolname can help their analysis and possible new direction for further development in the medical domain.
} % end of abstract

\begin{document}


\firstsection{Introduction}

\maketitle

Recent machine learning research on multimodal data has led to breakthroughs such as text-to-image generative models with extraordinary results (e.g., DALL$\cdot$E 2, Stable Diffusion, Midjourney). 
Another important advantage of multimodal models is that they can synthesize the information embedding in different modalities and lead to novel insights and better performance. 
Machine learning models can also benefit from the complementary information found in different modalities.
For example, variational autoencoders~\cite{AutoencodersHinton} are used to learn cross-modal representations in a multidimensional latent space, from large multimodal datasets~\cite{radhakrishnan_cross-modal_2023,MultidomainTranslationKarren}, which can provide state-of-the-art performances in various downstream predictive tasks.
Healthcare datasets like UK Biobank~\cite{UKBiobank} and MIMIC~\cite{johnson2020mimic} are excellent examples with which to learn clinically useful representations. 
In particular, cardiac magnetic resonance images (MRIs) are considered the gold standard for determining cardiac structure, while electrocardiograms (ECGs) are used ubiquitously for electrophysiological assessment~\cite{zipes2018braunwald}.
These complementary cardiac measurements can be used to provide clinical researchers with holistic insights into cardiovascular diseases.
% The \toolname uses a variational autoencoder to create a unified representation of MRI and ECG data. 
% and it can be generalized to other types of medical data or a different application domain. 
% To the best of our knowledge, there have been no published results on using autoencoders for multimodal medical data.

\begin{figure*}[th]
    \includegraphics[width=\linewidth]{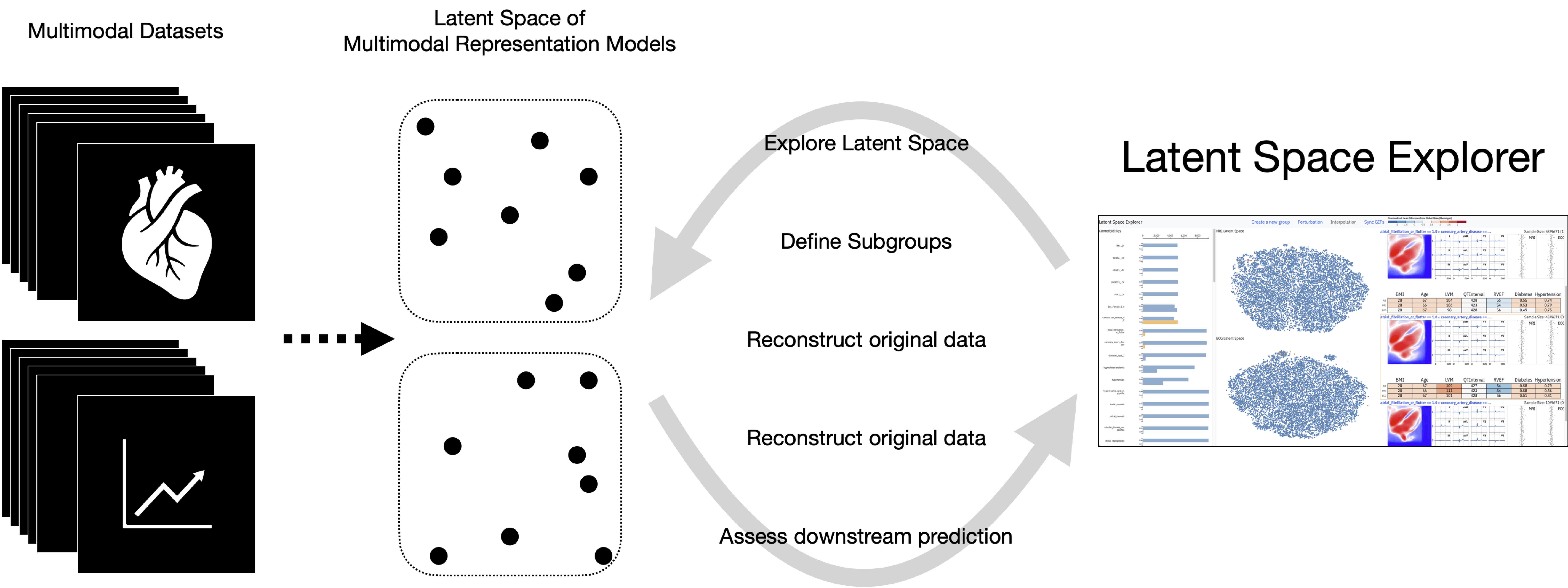}
    \caption{\toolname allows users to explore the latent space of multimodal representation models interactively.}
    \label{fig:mlsconcept}
    \vspace{-.65cm}
\end{figure*}

While such multimodal representation models can be powerful for downstream analysis, it is challenging for clinical researchers to explore the learned representations and inspect their clinical usefulness.
The learned representation is inherently difficult to interpret because it is so high-dimensional and must be assessed simultaneously for its reconstruction and downstream prediction capabilities.
Furthermore, existing visualizations are limited in their ability to help users like clinical researchers to understand the subject representations in the latent space and interactively predict various health-related outcomes. 
Furthermore, clinical researchers often desire to test the biases and fairnesses of the trained models on various subpopulations~\cite{kwon2022rmexplorer}.

% However, many machine learning models, not just multimodal ones, are difficult for domain experts to adopt because it is difficult to explore the models without visual aids and test the predictive performance of the models. 
% To address these challenges, we introduce a visual analytic system called \toolname that can synthesize the information from different modalities in a unified model and provide interactive visualizations that reduce the technical barrier for domain experts.
% In this paper, we use an example with cardiac MRIs and ECGs to demonstrate the benefits of \toolname. 
% but the approach can be extended to other types of medical data and even those from a different application domain.
% The use case demonstrates how clinical researchers can use multimodal models learned from two modalities to understand the status of multiple subjects with regard to cardiovascular diseases.
% In particular, cardiac magnetic resonance images (MRIs) are considered the gold standard for determining cardiac structure, while electrocardiograms (ECGs) are used ubiquitously for electrophysiological assessment~\cite{zipes2018braunwald}.
% These complementary cardiac measurements are often used together to give clinical researchers more holistic insights into heart health and pathology.

To address these challenges, we developed a visual analytics system called \toolname, which aims to help users to explore subject representations in the latent space learned from variational autoencoders on multimodal datasets.
Specifically, \toolname allows users to define subgroups of interest, interactively decode different modalities with the selected subjects, and inspect the accuracy of the embedding in downstream phenotype prediction tasks as Figure~\ref{fig:mlsconcept} depicts.
The design of \toolname is the result of collaboration with domain experts.
We demonstrate its usefulness with a usage scenario developed together with cardiology experts. The \toolname was evaluated by medical researchers on real-world datasets, and the results demonstrate its potential for more effective medical diagnosis and a new direction for further development.

\begin{table}[b!]
\centering
\small
\begin{tabular}{lrr}
\toprule
\multirow{ 2}{*}{Covariate}  &  Median or Total & Inter-quartile range \\
 & (N=37774) & or proportion (\%) \\
\midrule
           Age at Assessment &                      65 &                  (58-70) \\
                         BMI &                      25.98 &            (23.62,28.78) \\
          Genetic Sex Female &                   19496 &                   51.61\% \\
         Atrial Fibrillation &                    1335 &                    3.53\% \\
     Coronary Artery Disease &                    1330 &                    3.52\% \\
             Diabetes Type 2 &                    1586 &                    4.20\% \\
                Hypertension &                   11607 &                   30.73\% \\
 Hypertrophic Cardiomyopathy &                      34 &                    0.09\% \\
\bottomrule
\label{tab:table1}
\end{tabular}
\caption{Demographic and comorbidity data for the UK Biobank individuals with both ECGs and MRIs, used to train, evaluate, and visualize the cross-modal model.}
\vspace{-.5cm}
\end{table}

\section{Related Work}

In this section, we review prior work related to multimodal representation models and visual analytics applications to explore them.
\subsection{Multimodal Machine Learning}
Our approach relies on a class of machine learning models called autoencoders~\cite{BengioContractiveAutoencoders,DenoisingAutoencoders,AutoencodersBaldi}. Autoencoders are generative models which learn representations from unlabeled data by minimizing a reconstruction loss. A line of recent works uses autoencoders to learn joint representations of multi-modal data including natural images and captions \cite{DALLE, CLIP, ContrastiveCoding, ConstrastiveSimCLR, ContrastiveMoCoHe, ContrastivePIRLMisra}, gene expression in biology \cite{MultidomainTranslationKarren}, and paired clinical measurements \cite{PCLR, ContrastiveLearningDermatology, ContrastiveLearningMRI, radhakrishnan_cross-modal_2023}. The results show that autoencoders perform competitively with other multi-modal integration methods including classical integration approaches using canonical correlation analysis (CCA) \cite{Ballentine2022trips, DCCA, SeuratRef1, SeuratRef2} and generative adversarial networks (GANs) \cite{CycleGAN, MAGAN}.

\subsection{Visual Analytics for Latent Space Representation}

Many visual analytics tools were developed to explore and explain latent space representation for different applications.
% Computational + Visualization Approaches
Liu et al.~\cite{liu_towards_2020} proposed a method to visually explain variational autoencoders (VAEs) by visualizing a heatmap of gradient-based network attention extracted from the learned latent space representation. Embeddings extracted from hidden layers of neural networks can be used to explore visual concepts in image classification models~\cite{zhao_human_in_the_loop_2022, huang_conceptexplainer_2023}.
% Visualizations Approach
Various visual analytics approaches have been developed to explore, interact with, and compare embeddings in the latent space of machine learning models. SpaceSheets allows users to explore latent space through a spreadsheet user interface~\cite{loh_spacesheets_2018}. DeepVID helps users to understand the latent space by generating neighbors near an item of interest~\cite{wang_deepvid_2019}. Similarly, Embedding Comparator~\cite{boggust_embedding_2022} and Parallel embeddings~\cite{arendt_parallel_2020} enable visual comparison of local neighbors around each item in multiple embedding spaces. Furthermore, Emblaze~\cite{sivaraman_emblaze_2022} allows users to interactively explore embedding spaces on a visualization system integrated within an interactive computing environment. A visual analytics approach was adopted to explore the latent space model of motion data from elicitation studies~\cite{dang_gesturemap_2021}. Ross et al.~\cite{ross_evaluating_2021} proposed an interactive reconstruction task for inspecting the quality of interpretability of generative models.
Though these approaches provide inspiring techniques, the previous systems are not developed to explore latent space extracted from multi-modal, healthcare datasets or to help clinical researchers to gain clinical insights.

\section{\toolname}

In this section, we describe the design of \toolname. 
\toolname aims to help users explore the cross-modal latent space representation that includes cardio-vascular structural information and myoelectric information from MRIs and ECGs, respectively.
We also describe the dataset and the model we used while developing the system and the views of the system.

\begin{figure*}[t!]
    \includegraphics[width=\linewidth]{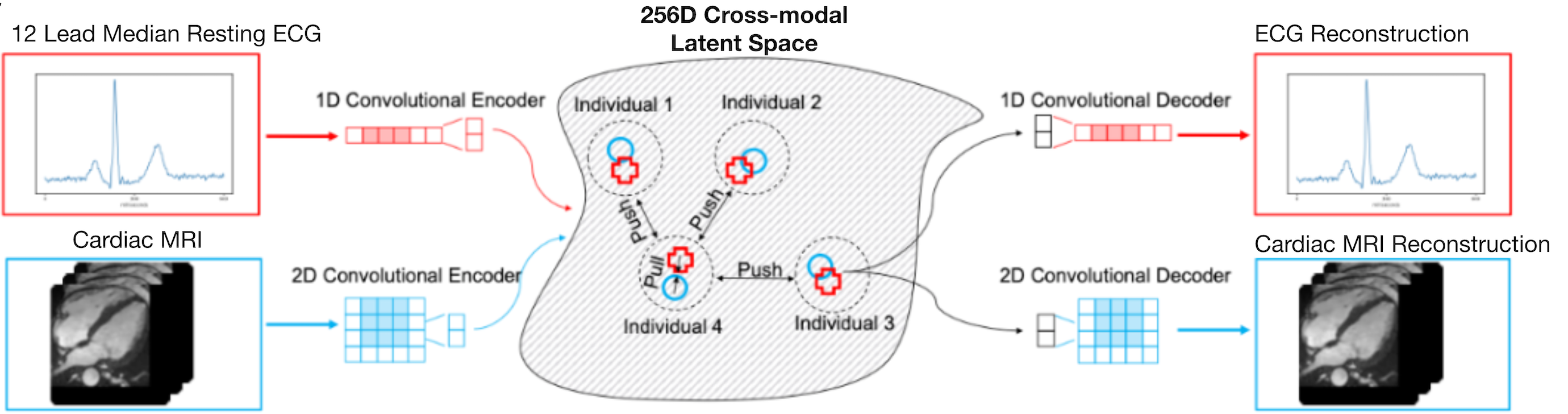}
    \caption{Cross-modal autoencoder framework uses reconstruction and contrastive loss to ensure different modalities from the same individual are embedded near each other, adapted from \cite{radhakrishnan_cross-modal_2023}.}
    \label{fig:autoencoder}
    \vspace{-.5cm}
\end{figure*}

\subsection{Requirements}

We used the medical diagnosis as an example domain and designed \toolname in collaboration with clinical researchers. 
We collaboratively derived the requirements and usage context. 
% While some of the requirements may appear domain-specific, our goal is always to make \toolname as domain agnostic as possible.

\begin{enumerate}[nosep,label=R\arabic*]
    \item There is a strong clinical need to explore the potential of synthesizing the MRI and ECG data, which could provide new insights and/or improve the performance of existing analyses.
    \item An intuitive interface is required to enable the exploration, analysis, and interpretation of the data and results for users with limited technical backgrounds.
    \item Besides the MRI and ECG data, related information, such as patient demographics and pre-existing conditions, is needed. This would allow the clinicians to discover possible links between the two, which is an important part of the analysis.
    \item The system must be responsive to support the interactive interrogation of the model and its results.
    \item The technical and black-box nature of the model makes it difficult for clinical experts to interact with such a system and interpret its results. The tool must provide additional features that would allow the target users to perform meaningful control and adjustments, and assist in the interpretation of the results.
%    \item Related to the black-box nature, there is also a possible distrust towards the results produced. The tool must provide means for the users to validate the results in ways they are familiar with.
\end{enumerate}

% (A list of requirements)
% \begin{enumerate}
%     \item who are the users?
%     \item what kind of analyses?
%     \item types of data to support?
%     \item any requirements about the autoencoder/model?
%     \item any requirements about the visualization?
% \end{enumerate}

%(Are there any requirements from literature?)

% \subsection{System architecture}
% (a system architecture diagram?)

\subsection{Data \& Model}
\label{sec:data_model}

We extracted 12-lead resting ECGs and 4-chamber long-axis cardiac MRI measurements conducted at the same imaging visit from 37,774 individuals in the UK Biobank. 
These individuals were randomly split into training (N=26,328), testing (N=3,807), and validation (N=7,639) sets (Table~\ref{tab:table1}).

To train the cross-modal autoencoder with multimodal self-supervision, the reconstruction loss is combined with a contrastive loss which ensures that paired ECG and MRI samples (from the same subject) are mapped to nearby points in the latent space, while discordant pairs (from different subjects) are pushed apart (Figure~\ref{fig:autoencoder}).
The reconstruction and contrastive losses are simultaneously minimized with the ADAM optimizer on a NVIDIA V-100 GPU for 12 hours and the weights with minimum validation loss modality-specific encoders and decoders are serialized with early-stopping.

Our model is trained on ECG and MRI pairs, but we consider the practically relevant setting in which only one modality is available. We observe that predictive models applied to our cross-modal representations generally outperform supervised deep learning models which are restricted to training on the subset of data with labels, and predictions from unimodal autoencoder representations. Embedding both modalities in the same latent space combined with the generative capability of the decoders allows for modality translation. \toolname allows users to compare these options and decide how many/which modality to use (more details in Section~\ref{sec:downstream}).

\subsection{Explore Latent Space \& Define Subgroups}

To meet user requirements, particularly R3, we designed \toolname so that it allows users to explore the latent space and define subgroups of interest using two visualizations: i) subgroup bar charts; ii) t-distributed stochastic neighbor embedding (t-SNE) scatter plots.
Once users launch \toolname with the underlying model and the dataset, the system displays bar charts that show the distribution of subjects over a set of variables (Figure~\ref{fig:teaser}~(A)). 
The variables include demographic characteristics (e.g., sex, age) and pre-existing conditions or diseases (e.g., diabetes, hypertension).
The system also plots the subjects on a scatter plot using the output of t-SNE (Figure~\ref{fig:teaser}~(B)).
The scatterplots show natural groupings of subjects in the latent space extracted from MRIs and ECGs each.

To allow interactive exploration, the bar charts and scatterplots are connected via two-way brushing \& linking to help users explore various groups defined by a set of variables.
% Figure~\ref{fig:explore_subgroups} illustrates how users can use the interactive feature. 
Users can click on multiple bars in a series to set and refine filters for the selected group of subjects.
The changes are reflected and shown on the scatterplot on the right side.
In addition, users can lasso-select groups of subjects in proximity on the scatterplots to refine the filters for the selected subjects.
Then, the changes are also shown in the bar charts.
During this process, users can understand the characteristics of the selected group of subjects with respect to demographic information and pre-existing conditions.

The original MRI image and ECG data can be shown when more details are needed for a patient (Figure~\ref{fig:teaser}C). The decoder in \toolname also allows the fast reconstruction of a representative MRI image and ECG data for a group of patients that the user selects. For example, a user can select a group of patients in the scatter plot (Figure~\ref{fig:teaser}B) and choose ``Create a new group'', the system plugs in the latent space representation vectors into the decoder, which then reconstructs MRIs and ECGs (Figure~\ref{fig:teaser}C).
\toolname either chooses a subject that is closest to the centroid of the group 
or constructs a new vector (mean, median, centroid). The autoencoder used in \toolname allows almost instantaneous construction, making it well-suited for interactive exploration (R4).
%The MRIs and ECGs can provide clinically useful insights into the selected subjects.
The MRIs and ECGs of multiple patients and groups can be displayed simultaneously to facilitate comparison (Figure~\ref{fig:teaser}C shows two such sets).

\begin{figure*}[t]
  \centering
% \fbox{
  \includegraphics[width=.8\linewidth]{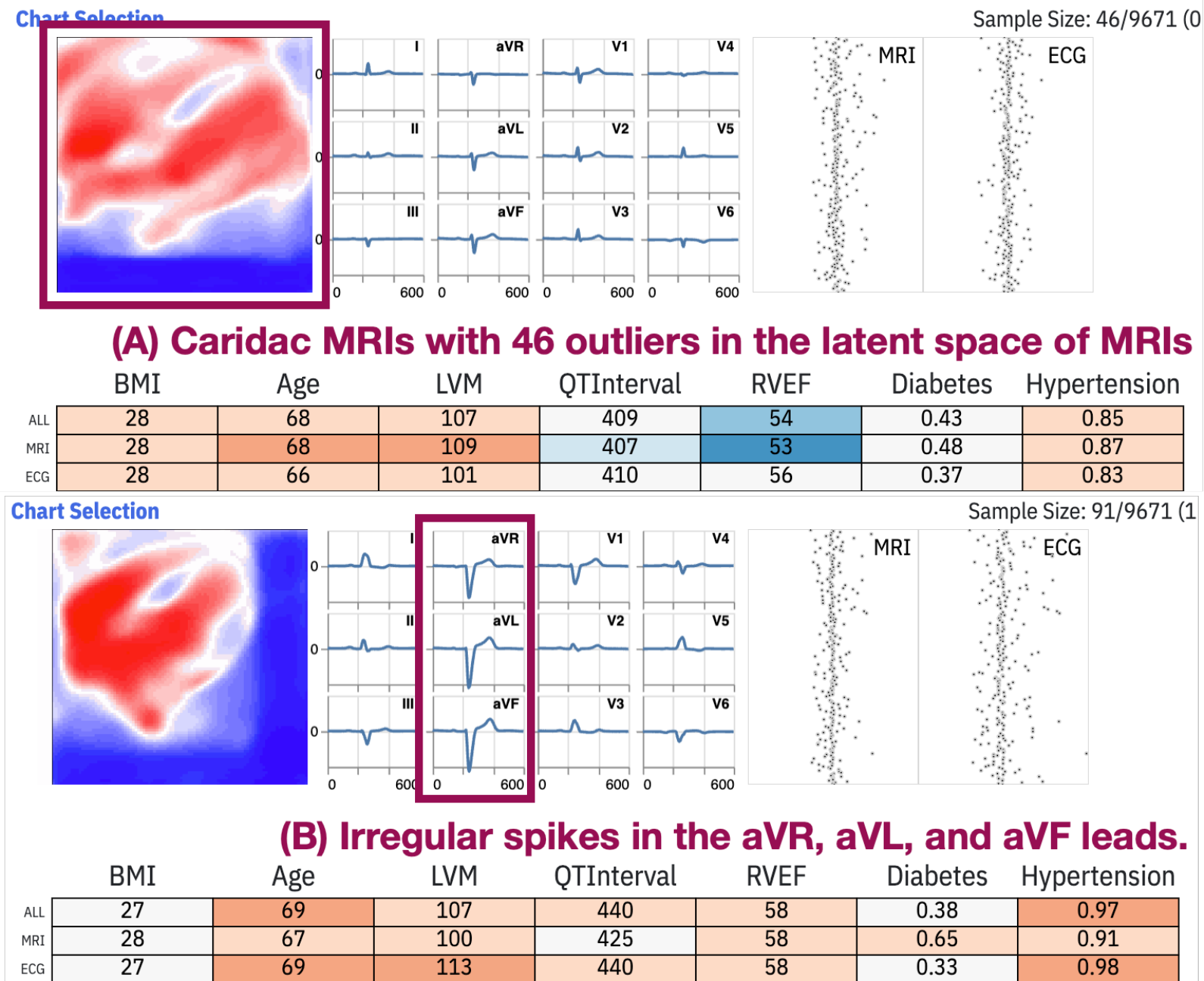}
% }
  \caption{Users can inspect the quality of images by interactively reconstructing MRIs and ECGs from selected groups of subjects. (A) MRIs show irregular visual features; (B) ECGs show abnormal spikes in the aVR, aVL, and aVF leads.}
  \label{fig:outliers}
  \vspace{-.65cm}
\end{figure*}

\subsection{Traverse Across Latent Space}

\toolname helps users to understand what each dimension of the trained latent space represents by allowing them to perturb the input latent space vectors and see how the change affects the reconstructed MRIs and ECGs (R5). 
Once a group of subjects is selected, users can choose ``Perturbation'' to open a dialog showing the reconstructed MRI, ECG, and input representation vectors in the latent space (Figure~\ref{fig:teaser}~(D)). 
The input representation vectors are shown as dot plots (v: dimension, h: value).
Each dimension can be adjusted interactively by moving its corresponding dot horizontally.
As soon as users make changes to a value of any dimension of latent space, the system reconstructs MRIs and ECGs by running the new vectors through the decoder. Again, the speed of the autoencoder in \toolname allows the result to be updated almost instantly (R4).
The newly generated MRIs and ECGs are shown next to the original ones so that users can compare them and understand how the changes users made in a latent space dimension can affect the visual characteristics of the MRIs and ECGs (R5).

\toolname also allows users to investigate the semantics captured in the latent space by traversing between two groups of subjects (Figure~\ref{fig:teaser}~(E)). Once Users create two groups of subjects by for example selecting them in the scatter plot (Figure~\ref{fig:teaser}~(B)), they can use ``Interpolate'' to open the traversing visualization ((Figure~\ref{fig:teaser}~(E)), with the two groups showing on the left and right respectively. 
Users can then adjust the slider between the two groups, which computes a linear interpolation of the two representation vectors of the two groups.
%in proportion to the distance they move the slider.
The newly generated vector is passed to the decoder to create MRIs and ECGs, which are shown in the middle so that users can inspect the differences in visual features of MRIs and ECGs in detail. This allows clinicians to explore the potential impact of certain patient features. For example, a clinician can select two sets of patients within different age groups (using the bar chart in (Figure~\ref{fig:teaser}~(A)), and explore the potential impact of age by interactively changing the interpolated representation.

\subsection{Investigate Downstream Predictive Performance}
\label{sec:downstream}

\toolname can be easily integrated into a larger medical analysis system. For example, it can be used to support downstream analysis such as 
%, users can probe the clinical usefulness of the latent space in 
predicting various health outcomes including comorbidities and phenotypes. In this case, 
users may have a classification/regression model that predicts the output variables of interest with the \toolname latent space as part of the input.
%The models trained with training data then can be used to infer outcomes with unseen test data.
Once integrated, \toolname can dynamically display the prediction results and compare the conditions of using single or multiple modalities. 
As Figure~\ref{fig:teaser}~(C) shows, once users create a group, the system shows the predicted variables using a heatmap table (at the bottom of the figure).
To compare the results of using single or multiple modalities, the heatmap has three rows showing predicted outcomes from ECGs only, MRIs only, and ECGs \& MRIs, over columns of predicted outcomes.
%By reading the heatmap, users can understand the usefulness of latent space for inferring subjects' health status.
This can help users choose whether to use single or multiple modalities and which modality(ies) to include.
The heatmaps can also be used to visualize the results of perturbed (Figure~\ref{fig:teaser}~(D)) or interpolated reconstructions (Figure~\ref{fig:teaser}~(E)).

\begin{figure*}[t]
  \centering
% \fbox{
  \includegraphics[width=.8\linewidth]{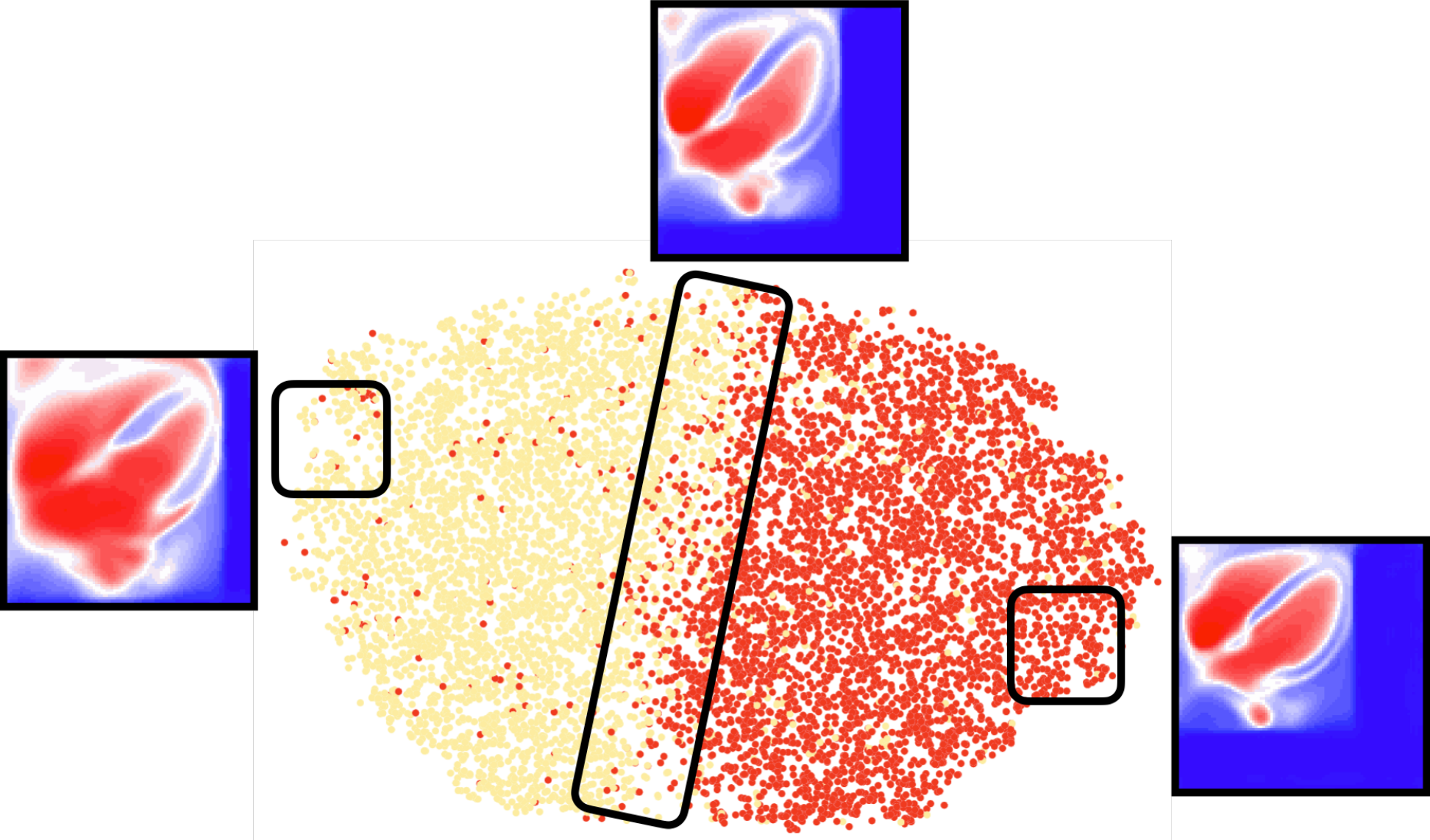}
% }
  \caption{Users can highlight scatter plots using any available demographic variable or phenotype. The figure highlights females in red and males in yellow. Using \toolname, users can reconstruct MRIs for groups selected in the scatter plots. For instance, the figure shows three reconstructed MRIs, from left to right: i) male only; ii) male and female on the boundary in the latent space; iii) female only.}
  \label{fig:gender_diffs}
  \vspace{-.65cm}
\end{figure*}

\section{Usage Scenario}
\label{sec:scenario}
In this section, we describe a usage scenario that 
%demonstrates how \toolname allows users to explore the multimodal latent space. 
was developed with cardiologists using the UK Biobank data discussed in Section~\ref{sec:data_model}.
%, who are co-authors of this paper.
% (Start with an overview of what the goal is and how it is achieved?)
% The scenario starts with inspecting and validating the collected data before moving on to the generation and testing of analysis hypotheses.

% Show: i) define subgroups; ii) explore neighbors in scatterplots; iii) visually reconstruct images using interpolation and extrapolation; iv) infer phenotypes by training models.
%% The noisy group: outliers
After loading the patient MRI and ECG data, cardiologist Emma 
%launches \toolname equipped with the trained cross-modal autoencoder, described in Section~\ref{sec:data_model}.
%She views the distribution of subjects over various phenotypes.
notices that 
%the cohort includes slightly more female than male subjects and a few subjects have various pre-existing medical conditions, such as diabetes, hypertension, and atrial fibrillation (Figure~\ref{fig:teaser}~(A)).
%After further examination of the individuals within the cohort using the scatter plot, she discovers 
a group of subjects who are distinct from the rest in the MRI plot (Figure~\ref{fig:teaser}~(B)).
To better understand its characteristics, Emma selects this group in the scatter plot using the lasso tool.
\toolname then reconstructs the representative MRI and ECG for this group (Figure~\ref{fig:outliers}~(A)). The reconstructed MRIs show irregular or unreadable visual features, which indicates that the scans failed to capture high-quality images.
Such failure is common, so Emma decides to drop this group from further analysis.
Similarly, she discovers another distinctive group within the ECG plot that has abnormal signals that are significantly below the baseline
%in the aVR, aVL, and aVF leads 
(Figure~\ref{fig:outliers}~(B)).
She decides to also discard these subjects because this is often caused by errors during data collection.

%% men vs women
After data cleaning by excluding possible inaccurate data, Emma turns the investigation to patient demographics, in this case, gender.
%differences between male and female subjects captured in the cross-modal latent space.
To do this, she first highlights the males and females in scatter plots (yellow: male, red: female) as shown in  Figure~\ref{fig:gender_diffs}.
The scatter plots divide the points into two halves: left with dominantly male subjects and right with dominantly female subjects.
To understand their differences, she reconstructs MRIs from three sampling groups: one female, one male, and one on the boundary.
As Figure~\ref{fig:gender_diffs} shows, the three reconstructed MRIs provide clear differences in their cardiac size between male subjects (large heart size) and female subjects (small heart size).
Those along the boundary tend to have an intermediate-sized heart. 
To confirm this, Emma activates the interpolation module 
%between the male and the female subjects 
(Figure~\ref{fig:teaser}~(E)).
Sliding from female to male subjects, she observes that interpolated samples have increasingly higher LVM (volume and size) and higher prevalence in diabetes and hypertension, which is very interesting.
%for male subjects in comparison to female subjects.

%% people with preexisting conditions (heart problems, diabetes, etc)
To test whether any pre-existing conditions can affect this, she creates a group of patients with a certain phenotype, namely hypertension and hypercholesterolemia.
She then further subdivides this group by gender.
The results show both similar differences, such as those in LVM and prevalence in type 2 diabetes, and new differences, such as those in prevalence in Right Ventricular Ejection Fraction (RVEF):
Males tend to have higher LVM but lower RVEF than females; 
Females, on the other hand, tend to have a lower prevalence of type 2 diabetes than males in the cohort.
This is consistent with existing literature that females tend to have a better right ventricular function than males~\cite{keen_sex_2021}
and the presence of diabetes is associated with right ventricular dysfunction~\cite{gorter_diabetes_2018}.
Based on these observations, Emma becomes more confident that \toolname together with the downstream analysis model can make accurate predictions.

%% annual household incomes

\section{User Evaluation}

We conducted a user evaluation with two domain experts to better understand how \toolname can help gain insights from the cross-modal representation derived from MRIs and ECGs.

\subsection{Methods}

% participants
We recruited two medical doctors who specialized in cardiology.
%Both of them are cardiology fellows, 
One of them was trained in a medical school in the US and the other in Germany.
Both of them have an introductory knowledge of machine learning (our targeted user type), as they were studying the application of machine learning techniques to understand the mechanism of various cardiovascular diseases
%(e.g., atrial fibrillation) using imaging, genetics, cell, and clinical data 
at the time of evaluation.
%They were invited to participate in our user evaluation session.

Pair analytics~\cite{pair-analytics} was used for the study, where one of the authors operated \toolname based on the instructions from the cardiologists. 
The Biobank data described earlier (Section~\ref{sec:data_model}) was used for the study and an exploratory analysis task, similar to the one described in the usage scenario (Section~\ref{sec:scenario}), was used in the study.
%and the two users provided comments via verbal comments.
The study was conducted remotely in an online video conference call and took approximately 1.5 hours.
%We explored the same dataset and model described in Section~\ref{sec:data_model} and discussed features of \toolname.
The analysis session was recorded, including the conversations between the Subject Matter Expert (SME, i.e., the cardiologist) and Visual Analytics Expert (VAE, i.e., the author). An interview was conducted after the analysis concluded to gather qualitative feedback.
% (what data is collected and how it is analysed.)
% dataset, scenario, method

\subsection{Results and Discussion}

Overall, the experts found \toolname intuitive and easy to understand. They commented that it would be impossible to use the underlying autoencoder without the interactive interface.
%praised the practicality of the visual analytics application.
They found it useful to explore independent cohorts defined by various clinical attributes and to generate cardiac MRIs and ECGs of them.
Also, they reported that it is helpful to show estimated phenotypes inferred from the latent space representation of patients.
During the session, they suggested various AI-assisted features that can be helpful to make the application even more powerful.
The most representative one is automatic segmentation for size and volumetric measurement of hearts from reconstructed images.
In particular, they wanted to find the correlation between volumetric, functional, and morphological features of hearts and the physiological status of patients. 
% In future work, we plan to research how to incorporate such features and what visual aids can help users perform such analyses.

% Need revision - here I directly copied my notes 
The evaluation results also showed that training is essential for such as tool, especially when cardiologists are to use it independently. Even with some preliminary machine learning knowledge, both cardiologists found it difficult to interpret the data in the scatter plot (Figure~\ref{fig:teaser}~(B)). We hypothesize that some understanding of the dimensionality reduction technique is required to interpret the data in a scatter plot and this can be potentially achieved through pre-training.
%Interestingly, clinical researchers did not find the "latent space” representations meaningful, especially in the TSNE view and the reconstruction view. 
%The TSNE views did not mean much to them because the groups of dots did not easily translate to clinical meanings.
%For the same reason, 
Similarly, the cardiologists found it difficult to use interactive perturbation (Figure~\ref{fig:teaser}~(D)) and interpolation (Figure~\ref{fig:teaser}~(E)). This is not surprising given that they require some knowledge about the underlying autoencoder model. 
We conjecture that even pre-train might not be sufficient to cover such a knowledge gap and such features may be better suited for data scientists and machine learning engineers who work alongside clinical researchers. 

Nonetheless, the cardiologists found these features interesting after they were explained. They would prefer a different interface, such as a 
%Rather, researchers envisioned a much easier 
natural language interface (e.g., ChatGPT), to control the generation of cardiac MRIs and ECGs of representative patients that satisfy certain conditions. 
%They wanted to express their queries with words instead of clicking comorbidities on the left view. 
%ChatGPT seemed to lower the barrier for “text-based” AI output generation. 
For instance, they wanted the system to take an input sentence like ``60-year old, male, previously diagnosed with atrial fibrillation,'' which can then be translated to reconstruct the representative MRIs and ECGs.
%In future work, researchers can train a text-to-image model and develop a system for such use.

Both cardiologists suggested that \toolname could be highly useful for medical training, which is a new use case we did not consider until then. 
They believed that both the instructors and students could use the tool to interactively reconstruct "representative" examples of cardiac MRIs and ECGs for cardiology training. 
For instance, an instructor can visually illustrate how LVMass of patients with prior conditions like heart failures differ from healthy hearts, or reconstruct the ECG irregularities of patients who are pregnant.
In addition, students can use \toolname to study the associations between patterns in MRIs and EKGs and patient outcomes like cardiomyopathy. 
%According to the experts, interactive reconstruction of medical images and their association with clinical outcomes seem useful features for teaching medical students--experienced cardiologists can run ``interactive dialogues'' with \toolname to teach medical students about how cardiovascular systems function differently for different subpopulations. 
Both cardiologists emphasized that the tool would be especially useful when real data cannot be used due to regulations and privacy concerns. 
%because the learned representation can generatively reconstruct images. 

One of the lessons for future multimodal latent space exploration tools design is to show multimodal information in multiple coordinated views.
Electronic medical records often capture subjects' characteristics using various modalities, such as imaging, text, and audio.
Various sources of information can enrich the understanding of the health status of subjects.
Using a representation learning approach to encode such highly complex information into a more compressed format can help us accelerate scientific discovery for pathology and physiology.
However, the multimodal representation can only be understood by providing the context in the original data formats.
\toolname provides multiple, coordinated views so that users can interactively explore the latent space and reconstruct MRIs and ECGs in order to understand the latent space.
Future work can investigate methods to show the connection between the latent space representation and the original data dimensions.

\section{Conclusion}

In this paper, we introduce a visual analytics system, \toolname, that is designed to support the interactive exploration of data with multiple modalities.
%It consists of multiple, coordinated views which help users to explore the multimodal representation of subjects, interactively reconstruct MRIs and ECGs with user-defined groups of subjects, and predict the phenotypes of the subjects using the latent space representation.
We demonstrated how \toolname can help reveal interesting relationships, such as the correlation between patient gender and prevalence of diabetes, and help assess the models' downstream predictive performance. 
The autoencoder approach provides opportunities for interactive visual exploration, and the ability to synthesize representative data on demand makes the tool particularly suitable for situations where data privacy is a major concern. 
We described a usage scenario developed with cardiologists using real-world Biobank data.
%demonstrating its usefulness for both ML practitioners and clinical researchers. 
A pair analytics evaluation was conducted with domain experts and provided evidence that \toolname largely achieves the original design goals and pointed out future research directions and new use cases for medical training.

%Future work can investigate how interactive visualization methods like ours can help clinicians and patients at the point of care.

%% If specified like this the section will be committed in preview mode
% The authors wish to thank A, B, and C. This work was supported in part by a grant from XYZ.}

%\bibliographystyle{abbrv}
\bibliographystyle{abbrv-doi}

\bibliography{lse}
\end{document}